\documentclass[twoside,11pt]{article}

\usepackage{jmlr2e}
\usepackage{amsmath}
\usepackage[usenames, dvipsnames]{color}
\usepackage{graphicx}
\usepackage{epstopdf}
\newcommand{\comment}[1]{}

\DeclareMathOperator*{\argmax}{argmax}

\usepackage[utf8]{inputenc}
\usepackage[english]{babel}
\usepackage{natbib}
\usepackage{bm}
\usepackage{amsmath}
\usepackage{multirow}
\usepackage{appendix}
\usepackage{dsfont}
\usepackage{appendix}

\makeatletter
\newcommand\appendix@section[1]{%
  \refstepcounter{section}%
  \orig@section*{Appendix \@Alph\c@section #1}%
  \addcontentsline{toc}{section}{Appendix \@Alph\c@section #1}%
}
\let\orig@section\section
\g@addto@macro\appendix{\let\section\appendix@section}
\makeatother

\ShortHeadings{Further results on structured regression for multi-scale networks}{Ba\v{s}i\'{c}, Arsi\'{c} and Obradovi\'{c}}
\firstpageno{1}

\begin{document}

\title{Further results on structured regression for multi-scale networks}

\author{\name Milan Ba\v{s}i\'{c} \email basic\_milan@yahoo.com \\
       \addr Department of Computer Science\\
       University of Ni\v{s}\\
       Vi\v{s}egradska 33, 18000 Ni\v{s}, Serbia
       \AND
       \name Branko Arsi\'{c} \email brankoarsic@kg.ac.rs \\
       \addr Department of Mathematics and Informatics\\
       University of Kragujevac\\
       Radoja Domanovi\'{c}a 12, 34000 Kragujevac, Serbia
       \AND
       \name Zoran Obradovi\'{c} \email zoran.obradovic@temple.edu \\
       \addr Department of Computer and Information Sciences\\
       Center for Data Analytics and Biomedical Informatics \\
       Temple University\\
       Philadelphia, PA, USA
}


\maketitle

\begin{abstract}
\comment{
%
}

Gaussian Conditional Random Fields (GCRF), as a structured regression model, is designed to achieve higher regression accuracy than unstructured predictors at the expense of execution time, taking into account the objects similarities and the outputs of unstructured predictors simultaneously. As most structural models, the GCRF model does not scale well with large networks. One of the approaches consists of performing calculations on factor graphs (if it is possible) rather than on the full graph, which is more computationally efficient. The Kronecker product of the graphs appears to be a natural choice for a graph decomposition. However, this idea is not straightforwardly applicable for GCRF, since characterizing a Laplacian spectrum of the Kronecker product of graphs, which GCRF is based on, from spectra of its factor graphs has remained an open problem. In this paper we apply new estimations for the Laplacian eigenvalues and eigenvectors, and achieve high prediction accuracy of the proposed models, while the computational complexity of the models, compared to the original GCRF model, is improved from $O(n_{1}^{3}n_{2}^{3})$ to $O(n_{1}^{3} + n_{2}^{3})$. Furthermore, we study the GCRF model with a non-Kronecker graph, where the model consists of finding the nearest Kronecker product of graph for an initial graph. Although the proposed models are more complex, they achieve high prediction accuracy too, while the execution time is still much better compare to the original GCRF model. The effectiveness of the proposed models is characterized on three types of random networks where the proposed models were consistently away more accurate than the previously presented GCRF model for multiscale networks [Jesse Glass and Zoran Obradovic. \emph{Structured regression on multiscale networks}. IEEE Intelligent Systems, 32(2):23-30, 2017.]. Moreover, the comparison of the GCRF models which use different approximations for the eigenvalues and eigenvectors is performed.
\end{abstract}


\section{Introduction}

Some real-life problems related to proteins-protein interactions, friendship on Internet social networks, traffic connections, Web pages and so on, could be considered as a graph-based problems. A graph (network) representations of the problems are specified as a set of objects which are connected among themselves. Since these relationships are application-specific, for a graph construction the prior knowledge about objects and relationship types among them should be known in advance, such as, relationships between documents can be quantified based on similarity of their contents \citep{radosavljevic2014neural}, relationships between pairs of scientific papers can be presented as the similarity of sequences of citation \citep{slivka2014distributed}, relationships between hospitals can be based on similarity of their specialization \citep{polychronopoulou2014hospital}, etc. There is also an example where the variety of interactions that exist among nodes are considered, e.g., the historical similarity of the two papers and the count of papers that cited both papers \citep{polychronopoulou2016multilayer}. Almost all real world networks evolve over time, either by adding or removing nodes or links over time. All of these processes often occur simultaneously, such as in social networks where users make and lose friends over time, thereby creating and destroying edges, and some users become part of new social networks or leave their networks, changing the nodes in the network. In these evolving networks, traditional predictive models, such as multivariate linear regression or neural networks, are necessary to extract missing information, identify spurious interactions, evaluate network evolving mechanisms, predict novel links and nodes attributes and so on. The unstructured regression models assume independent and identically distributed random variables and thus often fail to provide high accuracy in real-world applications that naturally have structured dependence. Unlike unstructured regression models, structured regression models such as Conditional Random Fields (CRFs) \citep{lafferty2001conditional}, the Markov Random Fields (MRF) \citep{solberg1996markov} and the Gaussian Conditional Random Fields (GCRF) \citep{radosavljevic2010continuous, qin2009global} are designed to incorporate the outputs of unstructured predictors and the correlation between objects in order to achieve higher regression accuracy. In other words, these models avoid independent and identically distributed random variables assumption by simultaneously learning to predict all outputs given all inputs. In this paper we deal with the Gaussian Conditional Random Fields (GCRF) model which allows the utilization of unstructured predictors as feature functions, and modeling of non-linear relationships between inputs and outputs. This model was first applied in computer vision \citep{liu2007learning}, but since then, there are vast research on different topics and applications \citep{polychronopoulou2014hospital, radosavljevic2010continuous, uversky2013links}, and model extensions for various purposes \citep{glass2016extending,gligorijevic2016uncertainty, stojkovic2016distance}.

In the Big Data era, some data are organized and stored as the large networks, that is why the efficient methods which are able to deal with such large amount of data are required. It turns out that the GCRF model is non-scalable for large networks with tens of thousands nodes. For example, a run time of the GCRF model for the network of order 100,000 nodes at each time step across more than 50 time points is approximately 2 months. UmGCRF \citep{glass2016extending}, the improved version of GCRF, takes more than one week and as such it is still not applicable for the real time decision-making (weather forecast, stock exchange etc.). Before the optimization algorithm of the UmGCRF model is started, a computationally inefficient operation for the large matrices is required and therefore the new improvements of the UmGCRF model are necessary.

One improvement of the UmGCRF model is possible to implement when a similarity matrix between outputs can be represented as the Kronecker product of graphs. This representation enables numerical calculations on its factors, instead on the entire matrix, thus the pre-processing operation executed before the learning task of the model becomes more efficient. The idea of exploiting the structure of Kronecker matrix multiplication is already studied. The benefits of the Kronecker product of graphs were successfully used in \citep{leskovec2010kronecker} where the authors fitted a Kronecker graph model into a real graph. In the paper \citep*{glass2017structured}, the fact that the whole network could be represented as the Kronecker product of graphs is used as a possibility for speeding up the GCRF learning task. Despite the idea of performing calculations on factor graphs rather than on the full graph is more computationally efficient, this idea is not straightforwardly applicable for the GCRF model. Indeed, in the GCRF model, Laplacian eigenvalues and their corresponding eigenvectors of the Kronecker product of graphs have to be calculated which appears as a new problem, since characterizing a Laplacian spectrum of such a graph from spectra of its factor graphs has remained an open problem. In this paper we apply new estimations of the Laplacian eigenvalues and the corresponding eigenvectors for the Kronecker product of graphs. By using suitable approximations for the eigenvalues and eigenvectors depending on certain type of networks, we significantly improve the regression accuracy in the same computational complexity time compared to the model presented in \citep*{glass2017structured}(mean squared errors of the approximated GCRF models are very close to the mean squared error of the original GCRF model). The conducted experiments over the three types of network (random, scale-free and small-world network) show that the proposed methods produce a reliable approximation for the eigenvalues and eigenvectors of the Laplacian matrix of the Kronecker product of graphs.Before the main results are presented, we give the analytical expressions of the approximations and a short overview of the difference between estimated and original eigenvalues of Laplacian of the Kronecker product of two graphs. The first part of the paper is devoted to the experiments where all estimated pairs of Laplacian eigenvalues and eigenvectors of the Kronecker product simulate the eigen-system (eigendecomposition) of the Laplacian matrix in the GCRF model. Here, we compare the loss in regression accuracy between the GCRF models which use these approximations and numerical calculations for the matrix eigendecomposition.

The second part of the paper is related to the non-Kronecker graphs. In the case when the network cannot be represented as a product of the networks such as Cartesian product, Kronecker product, strong product and so on, several approaches for speeding up the learning task of the GCRF model were proposed. In the method of \citep{ristovski2013continuous}, fully connected networks were considered in Euclidean feature space only. In order to make GCRF applicable to the large networks, \citep{zhou2016fast} proposed an approximation of the GCRF model by compressing a large (weighted and attributed) network into smaller one in such a way that the prediction accuracy on the reduced network is preserved. This model is based on the hypothesis that the compressed network maintains most information of the original network such that the loss in regression accuracy obtained by compressed GCRF is minor. A problem when the GCRF network can not be factorized as the product of networks is also considered in this paper. By applying Singular Value Decomposition algorithm, such a network can be approximated as the Kronecker product of two networks which allows us to reduce a given problem to the problem of speeding up the GCRF model which corresponds to the network that can be decomposed as the Kronecker product of two networks. The described representation enables us to apply obtained approximations for Laplacian spectrum and eigenvectors of the Kronecker product of graphs and this combined approach speeds up the GCRF model, while the loss in regression accuracy obtained by modified GCRF model is minor.

The paper is organized as follows. Section~\ref{sec:gcrf} gives an overview of the existing literature of significance for the GCRF prediction model. The GCRF model is determined by Laplacian of the similarity matrix $L(S)$, and in some cases $S$ can be represented as the Kronecker product of matrices which can be used for speeding up the learning task of the model. Since characterizing Laplacian spectrum of the Kronecker product of graphs from spectra of their factor graphs has remained an open problem, in Section~\ref{sec:gcrf-msn} we describe the reliable approximation methods for estimating the Laplacian eigendecomposition of the Kronecker product of graphs incorporated into the GCRF models. Section~\ref{sec:gcrf_performance} provides the details about the experimental setup, as well as experimental results of GCRF performances when the proposed estimations are applied. Finally, Section~\ref{sec:svd} discusses the results of the GCRF model when the similarity matrix can not be decomposed into the Kronecker product of matrices. In our approach two types of consecutively approximate methods are applied. The first approximation is used for finding the nearest Kronecker product of two matrices for an initial matrix, with regards to the Frobenius norm. Then, the approximation methods for the estimated Laplacian eigenvalues and eigenvectors of the Kronecker product of two graphs (described in Section~\ref{sec:gcrf-msn}) are applied on the obtained matrices as a necessary preprocessing step in the GCRF model. The paper concludes with a wrap up of key points and directions for further work.

\section{Background and related work}
\label{sec:gcrf}

Many real-life applications naturally have structured dependence which cannot be modeled with traditional unstructured predictive models. These unstructured models sometimes have strictly defined assumptions such as independent and identically distributed random variables and thus often provide low accuracy in learning tasks. Unlike unstructured learning, in structured learning, the model learns how to simultaneously predict all outputs given all input vectors by exploiting relationships that exist between multiple outputs. Mostly, those relationships are application-specific where the dependencies are defined in advance and as such can be represented by graphical models. In learning from spatial-temporal data, the Markov Random Fields \citep{solberg1996markov} and the more recently proposed Continuous Conditional Random Fields (CRF) \citep{qin2009global} are among the most popular graphical models. In this paper we will deal only with CRF model.

In CRF, as a type of the structured models, each of the $N$ feature vectors $\textbf{x}\in X \subseteq R^{d}$ (where $d$ is the number of features) interact with each of the outputs $y_{i}\in R$ through a mapping $f: X^{N}\rightarrow R^{N}$, while the outputs have influence on each other. These relationships between outputs express the conditional distribution between feature vectors and outputs from which we naturally obtain a representationally powerful graphical model and possibly improve accuracy. The conditional distribution $P(\textbf{y}|\textbf{x})$ for CRF can be represented in the following way

\begin{equation}
\notag
P(\textbf{y}|\textbf{x}) = \frac{1}{Z(\textbf{x}, \bm{\alpha}, \bm{\beta})} exp\Big(\sum_{i=1}^{N}A(\bm{\alpha}, y_{i}, \textbf{x}) + \sum_{j\sim i}I(\bm{\beta}, y_{i}, y_{j}, \textbf{x})\Big),
\end{equation}
where $A$ is an association potential with a $K$-dimensional parameter $\bm{\alpha}$, $I$ is an interaction potential with a $L$-dimensional parameter $\bm{\beta}$, and $Z(\textbf{x}, \bm{\alpha}, \bm{\beta})$ is a normalization function defined as

\begin{equation}
\notag
Z(\textbf{x}, \bm{\alpha}, \bm{\beta}) = \int\limits_{y} exp\Big(\sum_{i=1}^{N}A(\bm{\alpha}, y_{i}, \textbf{x}) + \sum_{j\sim i}I(\bm{\beta}, y_{i}, y_{j}, \textbf{x})\Big) dy.
\end{equation}
The purpose of the association potential $A$ is to represent relations between inputs and output in data, while interaction potential $I$ is to model interactions among outputs. In real-life applications, $A$ and $I$ are usually defined as a linear combination of a set of fixed feature functions $f_{k}$ and $g_{l}$, where $k=1,\ldots, K$ and $l=1,\ldots, L$, in terms of $\bm{\alpha}$ and $\bm{\beta}$ \citep{lafferty2001conditional}

\begin{equation}
\notag
\begin{gathered}
A(\bm{\alpha}, y_{i}, \textbf{x}) = \sum_{k=1}^{K}\alpha_{k}f_{k}(y_{i},\textbf{x})\\
I(\bm{\beta}, y_{i}, y_{j},\textbf{x}) = \sum_{l=1}^{L}\beta_{l}g_{l}(y_{i}, y_{j},\textbf{x}).
\end{gathered}
\end{equation}
Dominance of one of the potentials is reflected through the weights of the relevant feature functions, $\alpha_{k}$ and $\beta_{l}$, which will be determined during the learning process. The weights $\alpha_{k}$ and $\beta_{l}$ also determine the influence of the feature functions $f_{k}$ and $g_{l}$ within each of the potential, respectively.

If the feature functions are defined as quadratic functions of $\bm{y}$ \citep*{radosavljevic2010continuous}, the objective function $P(\textbf{y}|\textbf{x})$ becomes the probability density function of the multivariate Gaussian distribution where learning and inference tasks can be performed in a more efficient manner. This model is known as Gaussian Conditional Random Fields (GCRF). In this model, the association and interaction potential functions are defined as

\begin{equation}
\notag
A(\bm{\alpha}, y_{i}, \textbf{x}) = - \sum_{k=1}^{K}\alpha_{k}(y_{i}-R_{k}(\textbf{x}))^2,
\end{equation}
where $R_{k}(\textbf{x})$ is the $k$-th unstructured predictor (linear regression, neural network etc.) that predicts a single output $y_{i}$ taking into account $\textbf{x}$,
\begin{equation}
I(\bm{\beta}, y_{i}, y_{j},\textbf{x}) = - \sum_{l=1}^{L}\beta_{l}S_{i,j}^{l}(y_{i}-y_{j})^2,
\label{eq:interaction_potential}
\end{equation}
where $S^{l}$ represents a $l$-th similarity matrix between outputs $y_{i}$ and $y_{j}$. We can use as many matrices as we find necessary to model different similarity types between outputs. Now, the GCRF conditional probability is of the following form

\begin{equation}
\notag
P(\textbf{y}|\textbf{x}) = \frac{1}{Z(\textbf{x}, \bm{\alpha}, \bm{\beta)}} exp\Big(-\sum_{i=1}^{N}\sum_{k=1}^{K}\alpha_{k}(y_{i}-R_{k}(\textbf{x}))^{2} - \sum_{l=1}^{L}\sum_{j\sim i}\beta_{l}S_{ij}^{l}(y_{i} - y_{j})^2\Big).
\end{equation}
Such defined association and interaction potential enable GCRF to represent conditional probability form as a probability density function of multivariate Gaussian distribution

\begin{equation}
\notag
P(\textbf{y}|\textbf{x}) = \frac{1}{(2\pi)^{\frac{N}{2}}|\Sigma|^\frac{1}{2}}\, exp(-\frac{1}{2}(y-\mu)^T\Sigma^{-1}(y-\mu)),
\end{equation}
where $\Sigma^{-1}$ is the inverse covariance matrix

\begin{equation}
  \Sigma^{-1} = \left \{
  \begin{aligned}
    &2\sum_{k}\alpha_{k} + 2\sum_{h}\sum_{l}\beta_{l}S_{ih}^{l}, && \text{if}\ i=j \\
    &-2\sum_{l}\beta_{l}S_{ij}^{l}, && \text{if}\ i\neq j
  \end{aligned} \right.
  \label{eq:covariance_matrix}
\end{equation}
and $\mu$ is the expectation of the distribution

\begin{equation}
   	\mu = \Sigma b = 2\Sigma \sum_{k=1}^{K}\alpha_{k}R_k(\textbf{x}).
   	\label{eq:mean}
\end{equation}

For the optimization function $P(\textbf{y}|\textbf{x})$ and the given training set $\mathcal{D} = (X,\textbf{y}) = \\ \{(\textbf{x}_{i},y_{i})\}_{i=1,\ldots, N}$, the training task consists of estimation of parameters $\bm{\alpha}$ and $\bm{\beta}$ such that the conditional log-likelihood is maximized, $$\argmax_{\bm{\alpha}, \bm{\beta}} \sum_{y}logP(\textbf{y}|\textbf{x}).$$
To have a feasible model with real valued outputs, a normalization function $Z$ must be integrable. Discrete valued models are always feasible because $Z$ is finite and defined as a sum over finitely many possible values of $\bm{y}$.
The only remaining constraint is that $\Sigma^{-1}$ is positive semi-definite, which is a sufficient condition for the convexity of density function. One way to ensure that GCRF model is feasible is to impose the constraint that parameters $\bm{\alpha}$ and $\bm{\beta}$ are greater than 0. In this setting, learning is a constrained optimization problem. In order to satisfy these constraints and to convert the optimization problem to the unconstrained optimization, a technique from \citep{qin2009global} that applies the exponential transformation on $\bm{\alpha}$ and $\bm{\beta}$ parameters to guarantee that the new optimization problem becomes unconstrained is used.

Here, all parameters are learned by the gradient-based optimization. To be more precise, we will use the gradient descent algorithm. To apply it, we need to find the gradient of the conditional log-likelihood with first order derivatives

\begin{equation}
\begin{gathered}
\frac{\partial\, log\, P}{\partial\alpha_{i}} = \frac{-1}{2}(\textbf{y}^{T}\textbf{y} + 2R_{i}^{T}(\bm{\mu} - \textbf{y}) + \bm{\mu}^{T}\bm{\mu}) + \frac{1}{2}Tr(Q^{-1})\\
\frac{\partial\, log\, P}{\partial  \bm{\beta}} = \frac{-1}{2}(\textbf{y}^{T}L\textbf{y} + \bm{\mu}^{T}L\bm{\mu}) + \frac{1}{2}Tr(Q^{-1}L).
\end{gathered}
\label{eq:derivatives}
\end{equation}
where $Q = \frac{\Sigma^{-1}}{2}$. As shown in \citet{radosavljevic2010continuous}, the value that maximizes $P(\textbf{y}|\textbf{x})$ is equal to the mean $\mu$.

Beside the theoretical constraints, the main obstacle in determining optimization parameters of the GCRF model in real-life applications is computational complexity for the large networks. The gradient descent algorithm requires computing the inverse of the matrix $Q$ which takes $O(N^{3})$ time in each iteration. If the number of iteration is denoted with $I$, then the total running time for learning process is $O(IN^{3})$.

An improvement has come with the UmGCRF model \citep{glass2016extending}, the GCRF model for $l=1$ in \eqref{eq:interaction_potential}. Here, we shortly explain how they avoid expensive calculations of finding the matrix inverse inside the gradient descent algorithm by applying certain transformation on the matrix $Q$. From \eqref{eq:covariance_matrix} it is easy to see that the matrix $Q$ can be rewritten in the following way

\begin{equation}
\notag
Q = \sum_{k}\alpha_{k}I + \sum_{l}\beta_{l}L_{l},
\end{equation}
where $L_{l}$ is the Laplacian of the matrix $S^{l}$. For $l=1$, let $L_{1} = L = UDU^{T}$ be the eigendecomposition of the Laplacian matrix. Since $U$ is the orthonormal matrix then

\begin{equation}
Q = \sum_{k}\alpha_{k}I + \beta L = \sum_{k}\alpha_{k}UU^{T} + \beta UDU^{T} = U(\sum_{k}\alpha_{k}I + \beta D)U^{T}.
\label{eq:diagonalization}
\end{equation}
So, the eigenvalues of $Q$ are

\begin{equation}
\notag
\lambda_{i}=\sum_{k}\alpha_{k}+\beta d_{i}, \text{for all } i=1,\ldots, N.
\end{equation}
where $d_{i}$ are the diagonal elements of $D$, that is, the eigenvalues of $L$.
With certain pre-processing step which uses the eigenvectors $U$, the model learning task requires only a single eigendecomposition before the optimization algorithm is started, while the first order derivatives \eqref{eq:derivatives} can be computed in linear time since they are expressed in terms of scalar (eigenvalues $\lambda_{i}$), avoiding the matrix $Q$ inversion. The time complexity of this model is $O(N^{3}+IN)$. This model is significantly faster than the classical GCRF model, but still the model cannot handle large networks with more than several thousands of vertices. Also, the main drawback of the model is ability to work with only one similarity matrix defined between the outputs, that is for $l=1$ in \eqref{eq:interaction_potential}.

Although, structured regression in very large networks is often required, the previous approaches cannot handle them effectively. To address this problem \citet{glass2017structured} have taken advantages of the Kronecker product representation to speed up the GCRF learning task. This approach speed up the GCRF calculations in the case where the network structure can be represented as the Kronecker product of graphs. For that kind of matrices, the eigendecomposition of the Kronecker product of matrices can be reduced to the eigendecomposition of its factor matrices. The motivating task in their study was to predict monthly hospital admissions by disease by learning from millions of hospitalization records. The similarity matrix between these outputs can be represented as the Kronecker product of two networks, the network of 500 hospitals in the state of California and the network of more than 250 disease. However, characterizing Laplacian spectrum of the Kronecker product of graphs from spectra of their factor graphs has remained an open problem (there is not an explicit formula for this problem). Moreover, the Kronecker product of Laplacian matrices is not a Laplacian itself. For the application of their method in the GCRF model, the original Laplacian matrix is replaced with the normalized Laplacian matrix, since the normalized Laplacian behave well under Kronecker product, that is, its eigendecomposition can be determined easily. It should be also mentioned that very often the networks structure can not be represented as the Kronecker product of graphs. We will deal with this problem in Section~\ref{sec:svd}.

As the eigendecomposition in the prepocessing step has a large influence on final results of weights $\bm{\alpha}$ and $\bm{\beta}$ in the learning task, in most of the cases the eigenvectors have a larger influence than the network spectrum. Furthermore, more accurate eigenvector approximations give better eigenvalue approximation. The approximation proposed in \citep{sayama2016estimation} and our novel approximation give more relevant pairs of eigenvalue-eigenvector to the real ones, than the previous approach using only the eigendecomposition of the normalized Laplacian matrix. This statement will be confirmed with extensive experiments on random networks. Also, we will provide some theoretical evidences and show that the GCRF model which uses approximated eigenvalues and eigenvectors simultaneously outperforms the previous approach in synthetic datasets.

\section{Approximations for the eigenvalues and eigenvectors of the Kronecker product of graph}
\label{sec:gcrf-msn}
Before describing the proposed methods, we provide some notions and notations which will be used throughout the paper. A graph is a pair of sets $(V, E)$ where $V$ is a finite set called the set of vertices and $E$ is a set of 2-element subsets of $V$, called the set of edges. The adjacency matrix $A$ for a graph $G$ with $N$ vertices (nodes) is an $N\times N$ matrix whose $(i,j)$ entry is 1 if $(i,j)\in E$, and 0 otherwise. A number of the vertices $N$ of a graph $G$ is called an order of the graph $G$. If a weight $\omega_{ij}$ is assigned to each edge $(i,j)$ of the graph $G$, than $G$ is called weighted graph, that is, there exists a function $f:E\rightarrow R$. A generalization of the adjacency matrix $A$ is called a similarity matrix $S$, where $S_{ij}= \omega_{ij}$. The Laplacian matrix of the similarity matrix $S$ is defined as $L = D - S$ where $D$ is the degree matrix of $S$ (degree matrix is a diagonal matrix where each entry $(i,i)$ is equal to the sum of the weights of edges incident to $i$-th vertex). The normalized Laplacian matrix is defined as $\mathcal{L} = D^{-\frac{1}{2}}LD^{-\frac{1}{2}} = I - D^{-\frac{1}{2}}SD^{-\frac{1}{2}}$. Let $G = (V_{G}, E_{G})$ and $H = (V_{H}, E_{H})$ be two simple connected graphs, where $V_{G}$ ($V_{H}$) and $E_{G}\subseteq {V_{G}\choose 2}$ ($E_{H}\subseteq {V_{H}\choose 2}$) are the sets of vertices and edges of $G$ ($H$), respectively. The Kronecker product of graphs denoted by $G\otimes H$ is a graph defined on the set of vertices $V_{G}\times V_{H}$ such that two vertices $(g,h)$ and $(g',h')$ are adjacent if and only if $(g,g') \in E_{G}$ and $(h, h')\in E_{H}$. The Kronecker product of an $N\times N$ matrix $A$ and a $M\times M$ matrix $B$ is the $(NM)\times(NM)$ matrix $A\otimes B$ with elements defined by $(A\otimes B)_{I,J} = A_{i,j}B_{k,l}$ where $I = M(i-1)+k$ and $J = M(j-1) + l$. If $G$ and $H$ are weighted graphs, denote the similarity matrices of the graphs $G$ and $H$ by $S_{1}$ and $S_{2}$, respectively. The similarity matrix of the weighted Kronecker graph $G\otimes H$ is obtained as the Kronecker product of similarity matrices $S_{1}$ and $S_{2}$.

If we know that similarity matrix $S$ of $G\otimes H$ could be represented as the Kronecker product of similarity matrices of its factor graphs, then the matrix $Q$ from the GCRF model can be written in the following way

\begin{equation}
\label{eq:matrix_q}
Q = \sum_{k}\alpha_{k}I + \beta L_{S_{1} \otimes S_{2}}.
\end{equation}

As the Laplacian of the Kronecker product of graphs can not be represented in terms of its graph factors, we need to apply some of the approximations in order to obtain spectral decomposition of the Laplacian of the Kronecker product of graphs from those of its factor graphs.

\subsection{Estimation of Laplacian spectra of Kronecker product graph by using the Kronecker product of Laplacians eigenvectors}
\label{subsec:sayama_approx}

In the following text we will explain the motivation and assumptions from \citep{sayama2016estimation} for the proposed approximation. Laplacian of the Kronecker product of graphs is given by the following:
\begin{equation}
\begin{split}
L_{S_{1}\otimes S_{2}} & = D_{S_{1}\otimes S_{2}} - A_{S_{1}\otimes S_{2}}\\
& = (D_{S_{1}}\otimes D_{S_{2}}) - (A_{S_{1}}\otimes A_{S_{2}})\\
& = D_{S_{1}}\otimes D_{S_{2}} - (D_{S_{1}}-L_{S_{1}})\otimes(D_{S_{2}}-L_{S_{2}})\\
& = L_{S_{1}}\otimes D_{S_{2}} + D_{S_{1}}\otimes L_{S_{2}} - L_{S_{1}}\otimes L_{S_{2}},
\end{split}
\label{eq:sayama_start}
\end{equation}
where $A_{S_{1}}$ and $A_{S_{2}}$ are the similarity matrices and $D_{S_{1}}$ and $D_{S_{2}}$ are the degree matrices of graphs $S_{1}$ and $S_{2}$, respectively, where $|S_{1}| = n_{1}$ and $|S_{2}|=n_{2}$. The idea of the proposed approximation is to assume that $w_{i}^{S_{1}}\otimes w_{j}^{S_{2}}$, where $w_{i}^{S_{1}}$ and $w_{j}^{S_{2}}$ are arbitrary eigenvectors of $L_{S_{1}}$ and $L_{S_{2}}$ respectively, could be used as a substitute of the true eigenvectors of $L_{S_{1}\otimes S_{2}}$. Let $W_{S_{1}}$ and $W_{S_{2}}$ be $n_{1}\times n_{1}$ and $n_{2}\times n_{2}$ square matrices that contain all $w_{i}^{S_{1}}$ and $w_{j}^{S_{2}}$ as column vectors, respectively. Using \ref{eq:sayama_start} and by making another (mathematically incorrect) assumption that $D_{S_{1}}W_{S_{1}}\approx W_{S_{1}}D_{S_{1}}$ and $D_{S_{2}}W_{S_{2}}\approx W_{S_{2}}D_{S_{2}}$, after a short calculation it can be obtained

\begin{equation}
\begin{split}
L_{S_{1}\otimes S_{2}}(W_{S_{1}}\otimes W_{S_{2}}) & \approx (W_{S_{1}}\otimes W_{S_{2}})\Big(\Lambda_{S_{1}}\otimes D_{S_{2}} + D_{S_{1}}\otimes \Lambda_{S_{2}} - \Lambda_{S_{1}}\otimes \Lambda_{S_{2}}\Big)
\end{split}
\label{eq:laplacian_spectrum}
\end{equation}
where $\Lambda_{S_{1}}$ and $\Lambda_{S_{2}}$ are diagonal matrices with eigenvalues $\mu^{S_{1}}_{i}$ of $L_{S_{1}}$ and $\mu^{S_{2}}_{j}$ of $L_{S_{2}}$, respectively. From the last equation, estimated Laplacian spectrum of $S_{1}\otimes S_{2}$ could be calculated as

\begin{equation}
\mu_{ij} = \{\mu_{i}^{S_{1}}d_{j}^{S_{2}} + d_{i}^{S_{1}}\mu_{j}^{S_{2}} - \mu_{i}^{S_{1}}\mu_{j}^{S_{2}}\},
\label{eq:sayama_spectrum}
\end{equation}
where $d_{i}^{S_{1}}$ and $d_{j}^{S_{2}}$ are the diagonal entries of the degree matrices $D_{S_{1}}$ and $D_{S_{2}}$, respectively. When the eigenvalues are sorted in ascending order, the most effective heuristic method was observed. Now, we will explain how this effects the GCRF model.

According to the well-known property of the Kronecker product of decomposed matrices $(A\otimes B)(C\otimes D) = AC \otimes BD$, one can see that $(W_{S_{1}}\otimes W_{S_{2}})(W_{S_{1}}\otimes W_{S_{2}})^{T} = I$ since $W_{S_{1}}W_{S_{1}}^{T} = I$ and $W_{S_{2}}W_{S_{2}}^{T} = I$. If the matrix $(\Lambda_{S_{1}}\otimes D_{S_{2}} + D_{S_{1}}\otimes \Lambda_{S_{2}} - \Lambda_{S_{1}}\otimes \Lambda_{S_{2}})$ is denoted by $\mathcal{N}_{S_{1}, S_{2}}$, then the matrix $Q$ could be now rewritten as

\begin{equation}
\notag
Q = \sum_{k}\alpha_{k}I + \beta L(S_{1} \otimes S_{2}) \approx (W_{S_{1}}\otimes W_{S_{2}})(\sum_{k}\alpha_{k}I + \beta \mathcal{N}_{S_{1}, S_{2}})(W_{S_{1}}\otimes W_{S_{2}})^{T}.
\end{equation}

This approach reduces the model computational complexity because the calculations come down to smaller matrices, especially when the number of vertices is large. From \eqref{eq:laplacian_spectrum} and \eqref{eq:sayama_spectrum} it could be seen that the computational complexity of such a method is $O(n_{1}^{3} + n_{2}^{3} + n_{1}\, log\, n_{1} + n_{2}\, log\, n_{2} + n_{1}\, n_{2})$, where the cubic terms $n_{1}^{3}$ and $n_{2}^{3}$ represent the computational complexity of calculating Laplacian spectra of matrices $L_{S_{1}}$ and $L_{S_{2}}$, the terms $n_{1}\, log\, n_{1}$ and $n_{2}\, log\, n_{2}$ represent the computational complexity of sorting spectra and the eigenvalues multiplication costs $n_{1}\, n_{2}$. The complexity of explicit computation of eigenvalues and corresponding eigenvectors for $L_{S_{1}\otimes S_{2}}$ is $O(n_{1}^{3}n_{2}^{3})$ which is substantially larger than the complexity of the proposed approximation.

\subsection{Estimation of Laplacian spectra of Kronecker product graph by using the Kronecker product of normalized Laplacian eigenvectors}
\label{subsec:novel_approx}

Here, another approach for estimation of Laplacian spectra of the Kronecker product of graphs is described. The idea comes from the fact that the normalized Laplacian of the Kronecker product of graphs can be represented in terms of its factor graphs, more precisely, in terms of normalized Laplacian matrices of factor graphs. Moreover, in some
cases the Kronecker product of the eigenvectors of $\mathcal{L}_{S_{1}}$ and $\mathcal{L}_{S_{2}}$ gives better approximation for the eigenvectors of $L_{S_{1}\otimes S_{2}}$ than the Kronecker product of the eigenvectors of $L_{S_{1}}$ and $L_{S_{2}}$ [MATH\_PAPER]. By the definition of the normalized Laplacian and the properties $(A\otimes B)^{-1} = A^{-1}\otimes B^{-1}$ and $(A\otimes B)(C\otimes D)=(AC)\otimes (BD)$, the normalized Laplacian of the matrix $S_{1}\otimes S_{2}$ can be written in the following way

\begin{eqnarray*}
\mathcal{L}_{S_{1}\otimes S_{2}}&=& I_{n_{1}}\otimes I_{n_{2}} - (D_{S_{1}}^{-\frac{1}{2}} S_{1} D_{S_{1}}^{-\frac{1}{2}})\otimes (D_{S_{2}}^{-\frac{1}{2}} S_{2} D_{S_{2}}^{-\frac{1}{2}})\\
&=& I_{n_{1}}\otimes I_{n_{2}} - (D_{S_{1}}^{-\frac{1}{2}} S_{1} D_{S_{1}}^{-\frac{1}{2}})\otimes (D_{S_{2}}^{-\frac{1}{2}} S_{2} D_{S_{2}}^{-\frac{1}{2}}) \\
&=& I_{n_{1}}\otimes I_{n_{2}} - (I_{n_{1}} - \mathcal{L}_{S_{1}})\otimes (I_{n_{2}} - \mathcal{L}_{S_{2}}),
\end{eqnarray*}

\noindent where $D_{S_{1}}$ and $D_{S_{1}}$ are the degree matrices of the similarity matrices $S_{1}$ and $S_{2}$, respectively.

Let $\{\lambda_{i}^{S_{1}}\}$ and $\{\lambda_{j}^{S_{2}}\}$ be the eigenvalues of the terms $(I_{n_{1}} - \mathcal{L}_{S_{1}})$ and $(I_{n_{2}} - \mathcal{L}_{S_{2}})$, with the corresponding orthonormal eigenvectors $\{v_{i}^{S_{1}}\}$ and $\{v_{j}^{S_{2}}\}$, where $i = 1, 2, \ldots, n_{1}$ and $j = 1, 2, \ldots, n_{2}$. Denote by $\Lambda_{S_{1}}$ and $\Lambda_{S_{2}}$ the diagonal matrices whose diagonal elements are the eigenvalues $\lambda_{i}^{S_{1}}$ and $\lambda_{j}^{S_{2}}$, respectively. We also denote by $V_{S_{1}}$ and $V_{S_{2}}$ the square matrices which contain $v_{i}^{S_{1}}$ and $v_{j}^{S_{2}}$ as column vectors. Using the similar assumptions like in the previous subsection, in this case $D_{S_{1}}^{\frac{1}{2}} V_{S_{1}}\approx V_{S_{1}}D_{S_{1}}^{\frac{1}{2}}$ and $D_{S_{2}}^{\frac{1}{2}} V_{S_{2}}\approx V_{S_{2}}D_{S_{2}}^{\frac{1}{2}}$, after a short calculation we obtain the following formula

\begin{equation}
L_{S_{1}\otimes S_{2}}(V_{S_{1}}\otimes V_{S_{2}}) \approx (D\Lambda)(V_{S_{1}}\otimes V_{S_{2}}),
\label{eq:novel_spectrum}
\end{equation}

\noindent where $\Lambda = I_{n_{1}}\otimes I_{n_{2}} - \Lambda_{S_{1}}\otimes \Lambda_{S_{2}}$ and $D=D_{S_{1}}\otimes D_{S_{2}}$ (for more details see [MATH\_PAPER]). Since $D\Lambda$ in \eqref{eq:novel_spectrum} is a diagonal matrix, this leads us to a potential formula for the estimation of the Laplacian spectra of the Kronecker product of graphs

\begin{equation}
\notag
\mu_{ij} = \{(1-\lambda_{i}^{S_{1}}\lambda_{j}^{S_{2}})d_{i}^{S_{1}}d_{j}^{S_{2}}\}.
\end{equation}

This approach reduces the GCRF model computational complexity too, and the computational complexity of our approximation is $O(n_{1}^{3} + n_{2}^{3} + n_{1}^{2} + n_{2}^{2} + n_{1}\, n_{2})$. The computation complexity respect to the model from Subsection~\ref{subsec:sayama_approx} is similar, but there is a difference which is reflected in avoidance of spectra sorting, but we have matrix multiplication as an additional step in calculations $(I_{n_{1}} - \mathcal{L}_{S_{1}})$ and $(I_{n_{2}} - \mathcal{L}_{S_{2}})$. We notice that the same time complexity is obtained in the model from \citep*{glass2017structured}.

In the paper [MATH\_PAPER], a behavior of the estimated eigenvalues and eigenvectors of the presented approximations compared to the original ones is reported. All presented results were obtained in regard to the different types of graphs and different edge density levels. We will discuss in detail in Section~\ref{sec:gcrf_performance} how the estimated spectra and their corresponding eigenvectors influence the GCRF model. In order to experimentally characterize the GCRF model, experiments are performed on three types of graphs: Erd\H{o}s-R\'enyi, Barab\'asi-Albert and Watts-Strogatz, while the edge density percentage is varied over 10\%, 30\%, 50\%, 65\% and 80\%. For the orders of graphs $G$ and $H$ denoted by $n_{1}$ and $n_{2}$, respectively, we conduct all experiments three times depending on the orders of graphs $(n_{1}, n_{2})\in \{(30, 50),\- (50, 100),\- (100, 200)\}$.

\section{Performance of the GCRF model on synthetic networks}
\label{sec:gcrf_performance}


This section deals with the GCRF model performance check of different types of random networks under controlled conditions. The key role of the estimations, obtained in the previous section, is to overcome the computationally inefficient preprocessing step in the GCRF model, that is, Laplacian eigendecomposition of the large similarity matrix corresponding to the network obtained by the Kronecker product of networks. The used approximations have less computational complexity than the original GCRF model, but now, we need to check how the estimated eigenvectors and eigenvalues behave together and influence GCRF accuracy, having in mind that the GCRF model is very sensitive on the mathematical manipulations with the eigenvalues and associated eigenvectors.

The goal of this section is to determine the trade-off between the speedup in running time on one side, and the loss in regression accuracy on the other side, when the proposed estimations for the Laplacian spectra and eigenvectors of the Kronecker product of networks are applied. The reported results encompass different types of networks with a different number of vertices and edge density levels. The comparison is done between four models: \textit{GCRF-base} where the numerical eigendecompositon is performed and therefore the highest regression accuracy is achieved, and approximation models \textit{GCRF-MSN} \citep*{glass2017structured}, \textit{GCRF-LaplaceVec} (Subsection~\ref{subsec:sayama_approx}; \citet{sayama2016estimation}) and \textit{GCRF-NormLaplaceVec} (Subsection~\ref{subsec:novel_approx}) where the speed up of the learning task is in focus. Two general groups of experiments are performed for which we test

\begin{enumerate}
  \item \textbf{Model fitness:} The edge density level $\rho\in \{10\%, 30\%, 50\%, 65\%,\- 80\%\}$ is varied simultaneously for both graphs, $G$ and $H$, of given orders $n_{1}$ and $n_{2}$ where $(n_{1}, n_{2})\in \{(30, 50),\- (50, 100),\- (100, 200)\}$, and fixed noise sampled from $\mathcal{N}(0,\,0.33)$ is added to outputs.
  \item \textbf{Model robustness:} The noise sampled from $\mathcal{N}(0,\,0.25)$, $\mathcal{N}(0,\,0.33)$ and $\mathcal{N}(0,\,0.5)$ is added to outputs $Y_{train}$ in order to test the stability of all used approximations when the edge density level of the graphs $G$ and $H$ is fixed to 50\%, and the given graph orders are $(n_{1}, n_{2})\in \{(30, 50),\- (50, 100),\- (100, 200)\}$.
\end{enumerate}

Furthermore, for each general group of experiments, three types of random networks are used

\begin{itemize}
  \item \textbf{Random networks:} The first set of experiments is conducted on networks generated using the Erd\H{o}s-R\'enyi random network model.
  \item \textbf{Scale-free networks:} The second set of experiments is conducted on networks generated using the Barab\'asi-Albert network which reflects natural and human-made systems such as the Internet or social networks.
  \item \textbf{Small-world networks:} The third set of experiments is conducted on networks generated using the Watts-Strogatz network with small-world properties such as metabolic networks.
\end{itemize}

In the following text we describe a data-generation process of weighted, attributed, synthetic networks for experimental setup. In order to investigate the ability of the GCRF models which incorporate structure from various networks, we design the experiments where one type of random networks is used for each of the graphs, $G$ and $H$. First, two vectors $y_{1}$ and $y_{2}$ with the lengths $n_{1}$ and $n_{2}$ are generated from the normal distribution $\mathcal{N}(0,\,1)$. The vector of outputs $Y_{train}$, for the entire GCRF model, is generated as:
\begin{equation}
\notag
Y_{train} = y_{1}\otimes y_{2}+ \nu_{1}, \; \text{where}\; |y_{1}|=n_{1}, |y_{2}|=n_{2},\; \text{and}\; \nu_{1} \in \mathcal{N}(0, 0.33).
\end{equation}

\noindent The coordinates of the vectors $y_{1}$ and $y_{2}$ should be incorporated into the structure of the graphs $G$ and $H$, respectively, by assigning the certain weights to their edges thus creating dependence between the network structure and the model outputs. Therefore, the similarity matrix $S_{1}$ is obtained by assigning the weight $\omega(i,j) = e^{-(y_{1i}^{'}-y_{1j}^{'})}$ to the edge $(i,j)$ of the graph $G$, where a random noise is added to the vectors $y_{1}$ and $y_{2}$ i. e. $y_{1}^{'} = y_{1} + \nu_{2}$, $y_{2}^{'} = y_{2} + \nu_{2}$, and $\nu_{2}\in \mathcal{N}(0, 0.25)$, because the vectors should not be learned from the structure directly. The same holds for the similarity matrix $S_{2}$. After these steps, the structured (similarity) matrix $S$ for the GCRF model is obtained as the Kronecker product of the similarity matrices $S_{1}$ and $S_{2}$. A process of generation of unstructured predictor $R_{k}(x)$ ($k=1$) is done according to the equations \eqref{eq:covariance_matrix} and \eqref{eq:mean}, and $\alpha$ and $\beta$ are set to values 1 and 5, respectively. Later, this noise for $Y_{train}$ will be varied to test the robustness of the model to the noise. In the same way the test data are generated. The motivation for the added noises is to produce structured regression models which avoid overfitting caused by simultaneous learning of all given inputs to predict all outputs (the unstructured predictor $R_{k}(x)$ is directly obtain from $Y_{train}$). With added noise to $Y_{train}$, the Kronecker structure for $R_{k}(x)$ is avoided too. Therefore, we want to test whether the models provide good performance when there are small departures from parametric distributions.

For chosen parameters and for each of the GCRF models separately, we repeat these experiments independently one hundred times and the reported MSE (Mean square error) value is calculated as the average MSE value over the range from 5 to 95 percentiles. The confidence intervals for each of the models are calculated too. In the following experiments we show that the preprocessing errors caused by these approximations will slightly affect the loss in regression accuracy obtained by the GCRF model, when the considered networks are Erd\H{o}s-R\'enyi and Watts-Strogatz networks. Compared to these results, a gap between MSEs of the GCRF-base model on one side, and the used approximations on other side, is a bit higher for the Barab\'asi-Albert networks. All experiments were conducted on a PC with Intel Core i5-8265U 3.90 and 64 GB memory.

\subsection{Performance on Erd\H{o}s-R\'enyi and Watts-Strogatz random networks}

\begin{figure}
\centering
\includegraphics[width=\textwidth]{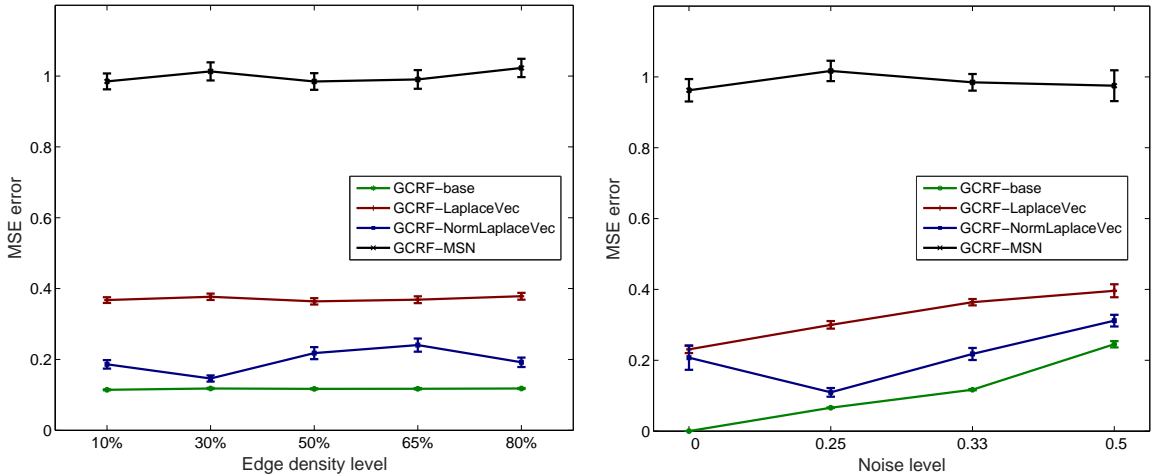}
\caption{Accuracy (MSE) of the models as a function of edge density percentage (left) and noise level (right), where $S_{1}$ is Erd\H{o}s-R\'enyi random network with 100 vertices and $S_{2}$ is Erd\H{o}s-R\'enyi random network with 200 vertices. \textit{Left:} the corresponding number of edges are \{122, 367, 612, 796, 980\} and  \{495, 1485, 2475, 3217, 3980\}. \textit{Right:} accuracy (MSE) of the models with respect to different noise, for fixed 50\% edge density percentage.}
\label{fig:er_100x200_density_noise}
\end{figure}

\textit{1) Model fitness (effectiveness with respect to edge density level):} Figure~\ref{fig:er_100x200_density_noise} (left panel) shows the prediction MSE of four approaches as a function of edge density percentage of the Kronecker product of two Erd\H{o}s-R\'enyi networks with 100 and 200 vertices. A fixed noise sampled from $\mathcal{N}(0,\,0.33)$ is added to outputs as well in order to completely conduct the model fitness checking experiment. The computational complexities of the used approximations are the same with respect to the term of the highest degree, but it is clear that GCRF-NormLaplaceVec produces more accurate regression results than both approaches, GCRF-LaplaceVec and GCRF-MSN. When the edge density level is set to 10\%, the regression MSE of the GCRF-LaplaceVec model is 0.37, that is almost two times higher value than the MSE of the GCRF-NormLaplaceVec model which is 0.19, while the GCRF-MSN has very high MSE. The same procedure is repeated for the networks with the same number of vertices, but with different edge density levels $\{30\%, 50\%, 65\%, 80\%\}$. When the edge density level is 30\%, GCRF-NormLaplaceVec achieves more accurate regression results than for all other edge densities. This can be explained by the fact that the eigenvalues corresponding to the eigenvectors $v_{i}^{S_{1}}\otimes v_{j}^{S_{2}}$ have the smallest distortion when the edge density level is exactly 30\% (see MATH\_PAPER). 

\begin{figure}
\centering
\includegraphics[width=\textwidth]{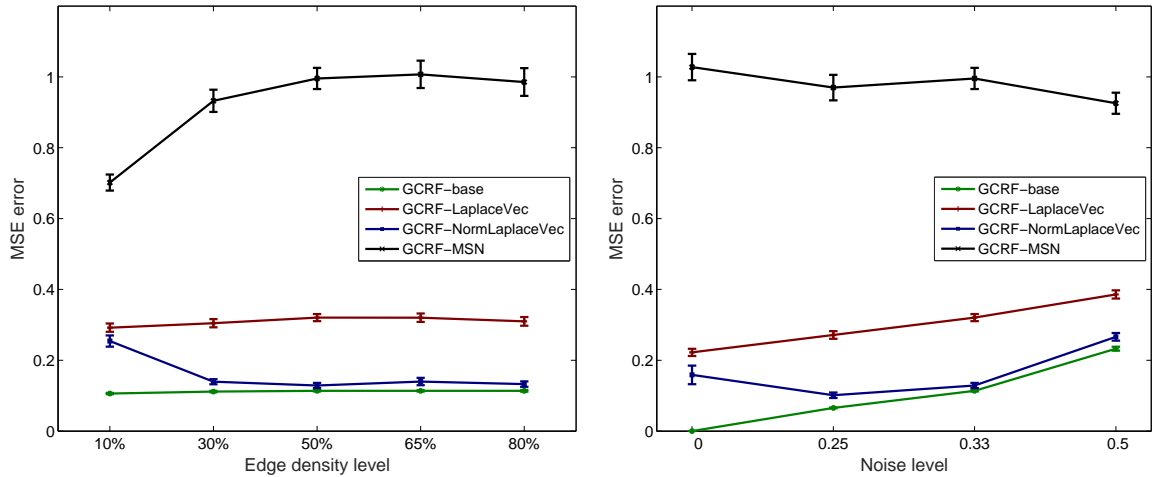}
\caption{Accuracy (MSE) of the models as a function of edge density percentage (left) and noise level (right), where $S_{1}$ is Watts-Strogatz random network with 100 vertices and $S_{2}$ is Watts-Strogatz random network with 200 vertices. \textit{Left:} corresponding number of edges are \{100, 350, 600, 800, 1000\} and  \{500, 1500, 2500, 3200, 4000\}. \textit{Right:} accuracy (MSE) of the models with respect to different noise, for fixed 50\% edge density percentage.}
\label{fig:ws_100x200_density_noise}
\end{figure}

The similar results are derived from Watts-Strogatz networks (left panel of Figure~\ref{fig:ws_100x200_density_noise}). When the edge density level is set to 10\%, the regression MSE of the GCRF-LaplaceVec and GCRF-NormLaplaceVec models are very close to each other, while the GCRF-MSN has the largest error. As the edge density grows, GCRF-LaplaceVec has almost the constant MSE at every point, while the GCRF-NormLaplaceVec error tends to the MSE of the GCRF-base model. At the same time, GCRF-MSN MSE gets close to 1.

High regression MSE of the GCRF-MSN model is always obtained, because the poor estimation of eigenvalues was used, although the eigenvectors are the same as the eigenvectors in the GCRF-NormLaplaceVec model. This can be explained with the fact that estimated eigenvalues in the GCRF-MSN model take values from the interval $[0,2]$, while the real eigenvalues belong to the interval $[0, n_{1}n_{2}]$ (this implies from the well known fact that all eigenvalues of the Laplacian matrix are in the range from 0 to the order of the matrix). The same experiments were also conducted on the Erd\H{o}s-R\'enyi networks with 30 and 50 vertices, as well as with 50 and 100 vertices, and the results are pretty much the same. For more details see Appendix~\ref{app:gcrf_small_graphs} and the left panels of Figures~\ref{fig:er_30x50_density_noise} and \ref{fig:er_50x100_density_noise}. The obtained MSEs have non-overlapping confidence intervals, thus the reported results are statistically significant.

\textit{2) Model robustness (effectiveness with respect to output noise):} In this group of experiments, for the fixed edge density level of 50\% for both networks, we vary the noise level in the model outputs in order to determine the robustness of the approximations against the noise in the vertex attributes. As it was expected, from the right panels of Figures~\ref{fig:er_100x200_density_noise} and \ref{fig:ws_100x200_density_noise} one can see that when the noise sampled from the Gaussian distributions $\mathcal{N}(0,\,0.25)$, $\mathcal{N}(0,\,0.33)$ or $\mathcal{N}(0,\,0.5)$ is added to the outputs, the accuracy performance of all models naturally decreases. MSE for GCRF-LaplaceVec and GCRF-NormLaplaceVec are almost the same when no noise is added to outputs in both cases, when Erd\H{o}s-R\'enyi and Watts-Strogatz random networks are used. But, when the noise is larger, the difference between MSEs becomes more noticeable in favor of the GCRF-NormLaplaceVec model, in both cases. It could be noticed that with a noise increasing, the GCRF-NormLaplaceVec error tends to the GCRF-base error. Again, MSE of the GCRF-MSN model is significantly higher compared to other approximations. The conclusion is pretty much the same for the smaller Erd\H{o}s-R\'enyi networks with 30 and 50 vertices, and 50 and 100 vertices (see right panels of Figures~\ref{fig:er_30x50_density_noise},~\ref{fig:er_50x100_density_noise} in Appendix~\ref{app:gcrf_small_graphs}).
To avoid repetition of the similar results, figures for the smaller Watts-Strogatz random networks are omitted.

The stability of the estimated eigenvalues and eigenvectors from Subsection~\ref{subsec:er_ws_vectors} is also reflected through the stability of the GCRF model. According to the results of the GCRF-NormLaplaceVec model on the Kronecker product of Erd\H{o}s-R\'enyi and Watts-Strogatz networks (see Figures~\ref{fig:er_100x200_density_noise} and \ref{fig:ws_100x200_density_noise}), which are very close to the results of the GCRF-base model, it turns out that the GCRF-NormLaplaceVec model is a very reliable approximation model.

\subsection{Performance on Barab\'asi-Albert networks}
\begin{figure}
\centering
\includegraphics[width=\textwidth]{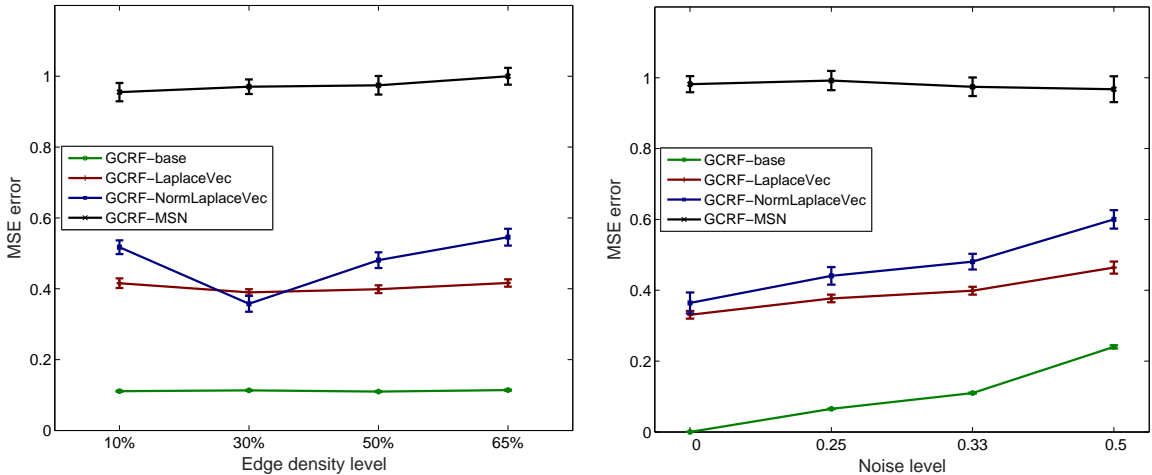}
\caption{Accuracy (MSE) of the models as a function of edge density percentage (left) and noise level (right), where $S_{1}$ is Barab\'asi-Albert random network with 100 vertices and $S_{2}$ is Barab\'asi-Albert random network with 200 vertices. \textit{Left:} corresponding number of edges are \{475, 1476, 2475, 2500\} and \{1900, 5904, 9900, 10 000\}. \textit{Right:} accuracy (MSE) of the models with respect to different noise, for fixed 50\% edge density percentage.}
\label{fig:ba_100x200_density_noise}
\end{figure}

The eigenvectors $w_{i}^{S_{1}}\otimes w_{j}^{S_{2}}$ and their corresponding eigenvalues are experimentally shown to be more suitable approximation for the original eigenvectors and eigenvalues for the Kronecker product of Barab\'asi-Albert networks than $v_{i}^{S_{1}}\otimes v_{j}^{S_{2}}$ and their corresponding approximated eigenvalues [MATH\_PAPER]. Therefore, it is expected that the GCRF-LaplaceVec model provides the highest regression accuracy in most of the cases and this is confirmed by conducting the experiments for two Barab\'asi-Albert networks which have 100 and 200 vertices, respectively. The results are presented in Figures~\ref{fig:ba_100x200_density_noise}. GCRF-LaplaceVec produces more accurate regression results than the other two approximations regardless of the edge density level and the noise added to the outputs, with exception when the edge density level is 30\%. We may notice that the smallest distortion of the eigenvalues corresponding to the eigenvectors $v_{i}^{S_{1}}\otimes v_{j}^{S_{2}}$ is when the edge density level is exactly 30\% (see Figure~\ref{fig:ba_eigenvalues_30x50}, MATH\_PAPER). For the smaller networks, the results could be seen in Figures \ref{fig:ba_30x50_density_noise} and \ref{fig:ba_50x100_density_noise} in Appendix~\ref{app:gcrf_small_graphs}.

We would like to point out that the gap between MSEs of the GCRF-NormLaplaceVec and GCRF-LaplaceVec models on one side, and MSE of the GCRF-base model on other side, is much higher for Barab\'asi-Albert networks compared to the Erd\H{o}s-R\'enyi and Watts-Strogatz random networks. We do not consider the GCRF-MSN model, since its MSE is always high. In the case of Barab\'asi-Albert networks, the smallest gap is around 0.3 considering all mentioned approximation models, while in the case of Erd\H{o}s-R\'enyi and Watts-Strogatz random networks is much less, almost 0 (around 0.028 and 0.014, respectively) (see left panels of Figures~ \ref{fig:er_100x200_density_noise}, \ref{fig:ws_100x200_density_noise} and \ref{fig:ba_100x200_density_noise}). Also, it can be noticed that in the case of Erd\H{o}s-R\'enyi and Watts-Strogatz networks the MSEs of the GCRF-NormLaplaceVec and GCRF-LaplaceVec models tend to MSE of the GCRF-base model (see right panels of the same figures), so in this case both models can be treated as the satisfactory ones. It seems that a task of improving the approximation models reliability is possible, so seeking new approximations could be new challenging direction in the future research that would establish lower MSE in the GCRF model.

\subsection{GCRF execution time with approximations}
\label{subsec:exec_time}

We proceed to compare the performance of the GCRF model when three approximations are used with respect to the GCRF-base model. The computational complexity for each approximation was individually explained in the previous sections, but we also check the execution time of the GCRF models, separately. Table~\ref{tbl:speed} shows the results when two networks are Erd\H{o}s-R\'enyi networks with the edge density level of 30\%. We also test the larger networks than ones used in the previous sections. Since GCRF-LaplaceVec has the same computational complexity compared to the GCRF-NormLaplaceVec and GCRF-MSN models, one can see that there is no significant difference in the speed between them. Moreover, since the GCRF models depend on the number of iterations, it can be seen that the GCRF-NormLaplaceVec and GCRF-MSN models have less iterations in the learning task than GCRF-LaplaceVec has. The baseline algorithm has large computational complexity of $O(n_{1}^{3}n_{2}^{3})$, and so it is slower in respect with the GCRF models with approximations, although its number of iterations decreases as the orders of networks increase. Therefore, a trade-off between execution time and regression accuracy of the GCRF models with approximation is accomplished in the case when the GCRF-NormLaplaceVec model is used for the random and small-world networks, and the GCRF-LaplaceVec model in the case of the scale-free networks. It should be mentioned that there is no significant difference in the execution time for Watts-Strogatz and Barab\'asi-Albert networks.

\begin{table}
  \footnotesize
  \centering
  \setlength{\tabcolsep}{0.5em}
  \renewcommand{\arraystretch}{1.2}
  \begin{tabular}{|c|c|c|c|c|c|c|c|c|c|}
      \cline{3-10}
      \multicolumn{2}{c}{} & \multicolumn{8}{|c|}{\textbf{GCRF}}\\
    \hline
      \multicolumn{2}{|c|}{\textbf{\#vertices}} & \multicolumn{2}{c|}{\textbf{base}} & \multicolumn{2}{c|}{\textbf{LaplaceVec}} & \multicolumn{2}{c|}{\textbf{NormLaplaceVec}} & \multicolumn{2}{c|}{\textbf{MSN}} \\
    \hline
    \textbf{G} & \textbf{H} & \textbf{ex.time} & \textbf{\#iter} & \textbf{ex.time} & \textbf{\#iter} & \textbf{ex.time} & \textbf{\#iter} & \textbf{ex.time} & \textbf{\#iter}\\
    \hline
    50 & 30 & 0.78 $\pm$ 0.03 & 21 & 0.31 $\pm$ 0.03 & 21 & 0.47 $\pm$ 0.01 & 16 & 0.47 $\pm$ 0.01 & 16\\ \hline
    50 & 100 & 19.03 $\pm$ 0.09 & 22 & 1.59 $\pm$ 0.04 & 22 & 1.74 $\pm$ 0.02 & 17 & 1.73 $\pm$ 0.01 & 20\\ \hline
    100 & 200 & 1089 $\pm$ 8.99 & 46 & 45.02 $\pm$ 0.54 & 47 & 44.88 $\pm$ 0.69 & 24 & 45.15 $\pm$ 0.49 & 22 \\ \hline
    100 & 300 & 2216 $\pm$ 6.21 & 9 & 82.25 $\pm$ 0.37 & 51 & 81.48 $\pm$ 0.58 & 24 & 81.09 $\pm$ 0.37 & 23 \\ \hline
    200 & 200 & 5192 $\pm$ 43.6 & 9 & 186.42 $\pm$ 2.46 & 73 & 185.13 $\pm$ 2.93 & 26 & 181.77 $\pm$ 2.25 & 26 \\ \hline
  \end{tabular}
  \caption{Execution time in seconds when the networks are Erd\H{o}s-R\'enyi networks with the edge density level of 30\%.}
  \label{tbl:speed}
\end{table}

\section{Performance of the GCRF model by using the Kronecker decomposition}
\label{sec:svd}

In the previous section we showed how the proposed approximations of the Laplacian spectra of the Kronecker product of graphs influence the GCRF model regression accuracy. These approximations are very suitable in the case when the similarity matrix $S$ in the GCRF model can be represented as the Kronecker product of the smaller similarity matrices which correspond to certain types of random graphs. In real-life applications, very often this is not the case. In this section we test the regression accuracy of the GCRF model, when the similarity matrix can not be decomposed into the Kronecker product of matrices. Our approach consists of two types of consecutively applied approximate methods, the first one consists of finding the nearest Kronecker product of the matrices of the given orders $S_{1}$ and $S_{2}$ to the similarity matrix $S$ and thereafter estimating the spectrum of $L(S_{1}\otimes S_{2})$ given the spectra of $S_{1}$ and $S_{2}$, respectively.

\subsection{Theoretical background}
\label{subsec:svd_theory}
First, we describe the algorithm for finding the nearest Kronecker product of matrices for an initial matrix $A$, with regards to the Frobenius norm defined as the square root of the sum of the absolute squares of the matrix elements. More precisely, for a given matrix $A\in \mathbb{R}^{m\times n}$, where $m=m_{1}m_{2}$ and $n=n_{1}n_{2}$, our task is to determine the matrices $B \in \mathbb{R}^{m_{1}\times n_{1}}$ and $C \in \mathbb{R}^{m_{2}\times n_{2}}$ such that $\Phi_{A}(B,C) = \parallel A - B\otimes C \parallel_{F}$ is minimized. This problem, also known as the nearest Kronecker product problem (\citet*{van1993approximation}), can be solved by using the singular value decomposition of a so called permuted matrix of $A$, denoted by $\mathcal{R}(A)$. In the following we give a precise definition of $\mathcal{R}(A)$ and a short overview of the solution for this optimization problem.

Consider the $m_{2}\times n_{2}$ submatrices (blocks) of the matrix $A\in \mathbb{R}^{m\times n}$,

\begin{equation}
\label{eq:blocking}
\begin{bmatrix}
    A_{11} & A_{12} & \dots & A_{1,n_{1}} \\
    A_{21} & A_{22} & \dots & A_{2,n_{1}} \\
    \vdots & \vdots & \ddots & \vdots \\
    A_{m_{1},1} & A_{m_{1},2} & \dots & A_{m_{1},n_{1}}
\end{bmatrix},\quad  A_{ij}\in \mathbb{R}^{m_{2}\times n_{2}},
\end{equation}
and the operation $vec: \mathbb{R}^{p\times q}\rightarrow \mathbb{R}^{pq\times 1}$ obtained by stacking the columns $X_{1,i}\in \mathbb{R}^{p\times 1}$, $1\leq i \leq q$, of a matrix $X$ on top of one another

\begin{equation}
\notag
vec(X)=
\begin{bmatrix}
    X_{1,1}\\
    X_{1,2}\\
    \vdots\\
    X_{1,q}
\end{bmatrix} \in \mathbb{R}^{pq\times 1}, X \in \mathbb{R}^{p\times q}.
\end{equation}
This operation will be used to express the minimization of $\parallel A - B\otimes C \parallel_{F}$ in terms of so-called a rank-1 approximation problem (for a given matrix $M$, a matrix $\widehat{M}$ with $rank(\widehat{M})=1$ should be determined such that $\parallel M - \widehat{M}\parallel_{F}$ is minimal). Furthermore, with respect to the blocks $A_{ij}\in \mathbb{R}^{m_{2}\times n_{2}}$ ($i=1,\ldots,m_{1}$, $j=1,\ldots,n_{1}$) of the matrix $A\in \mathbb{R}^{m\times n}$ from \eqref{eq:blocking}, where $m=m_{1}m_{2}$ and $n=n_{1}n_{2}$, the operator $\mathcal{R}: \mathbb{R}^{m_{1}m_{2}\times n_{1}n_{2}}\rightarrow \mathbb{R}^{m_{1}n_{1}\times m_{2}n_{2}}$ is defined in the following way

\begin{equation}
\notag
\mathcal{R}(A)=\begin{bmatrix}
    A_{1}\\
    A_{2}\\
    \vdots\\
    A_{n_{1}}
\end{bmatrix},\quad  A_{j}= \begin{bmatrix}
    vec(A_{1,j})^{T}\\
    vec(A_{2,j})^{T}\\
    \vdots\\
    vec(A_{m_{1},j})^{T}
\end{bmatrix},\quad  1\leq j\leq n_{1}.
\end{equation}
The following theorem establishes a connection between the problem of minimizing $\Phi_{A}(B,C)$ and the problem of approximating $\mathcal{R}(A)$ with a rank-1 matrix.

\begin{theorem}
\label{thm:approx}
Assume that $A \in \mathbb{R}^{m\times n}$ with $m=m_{1}m_{2}$ and $n=n_{1}n_{2}$. If $B \in \mathbb{R}^{m_{1}\times n_{1}}$ and $C \in \mathbb{R}^{m_{2}\times n_{2}}$, then
$\Phi_{A}(B,C) = \parallel A - B\otimes C \parallel_{F} = \parallel \mathcal{R}(A)-vec(B)vec(C)^{T}\parallel_{F}$.
\end{theorem}
The act of minimizing $\Phi$ is equivalent to finding a nearest rank-1 matrix of $\mathcal{R}(A)$. The approximation of a given matrix by a rank-1 matrix has a well-known solution, obtained from Theorem~\ref{thm:approx}, in terms of the singular value decomposition.

\begin{corollary}
\label{cor:svd}
Assume that $A \in \mathbb{R}^{m\times n}$ with $m=m_{1}m_{2}$ and $n=n_{1}n_{2}$. If $\tilde{A} = \mathcal{R(A)}$ has singular value decomposition
$$U^{T}\tilde{A}V = \Sigma = diag(\sigma_{i})$$ where $\sigma_{1}$ is the largest singular value, and $U(i,1)$ and $V(j,1)$ are the corresponding singular vectors $(i=1,\ldots,m_{1}n_{1}, j=1,\ldots,m_{2}n_{2})$, then the matrices $B \in \mathbb{R}^{m_{1}\times n_{1}}$ and $C \in \mathbb{R}^{m_{2}\times n_{2}}$ defined by $vec(B)=\sigma_{1}U(i,1)$ and $vec(C)=V(j,1)$ minimize $\parallel A - B\otimes C \parallel_{F}$.
\end{corollary}

In the following text we describe a data-generation process of weighted, attributed, synthetic networks and GCRF model parameters for experimental setup.

\subsection{Experimental setup}
\label{subsec:svd_exsetup}

In the rest of the section we present the obtained results for the GCRF models when the non-Kronecker similarity matrix is decomposed using the singular value decomposition according to the results from Corollary~\ref{cor:svd}. First, we briefly describe the experimental setup used in this section, which is slightly different from the setup described in the previous section. The vector of outputs $Y_{train}$, for the entire GCRF model, is generated as

\begin{equation}
\notag
Y_{train} = y_{1}\otimes y_{2}+ \nu_{1}, \; \text{where}\; |y_{1}|=n_{1}, |y_{2}|=n_{2},\; \text{and}\; y_{1}, y_{2}\in \mathcal{N}(0,1), \nu_{1} \in \mathcal{N}(0, 0.33),
\end{equation}

\noindent where the high correlation between the outputs and unstructured predictors $R_{k}(x)$ is removed by adding extra noise to $Y_{train}$, sampled from $\mathcal{N}(0, 0.33)$.

\begin{figure}
\centering
\includegraphics[width=\textwidth]{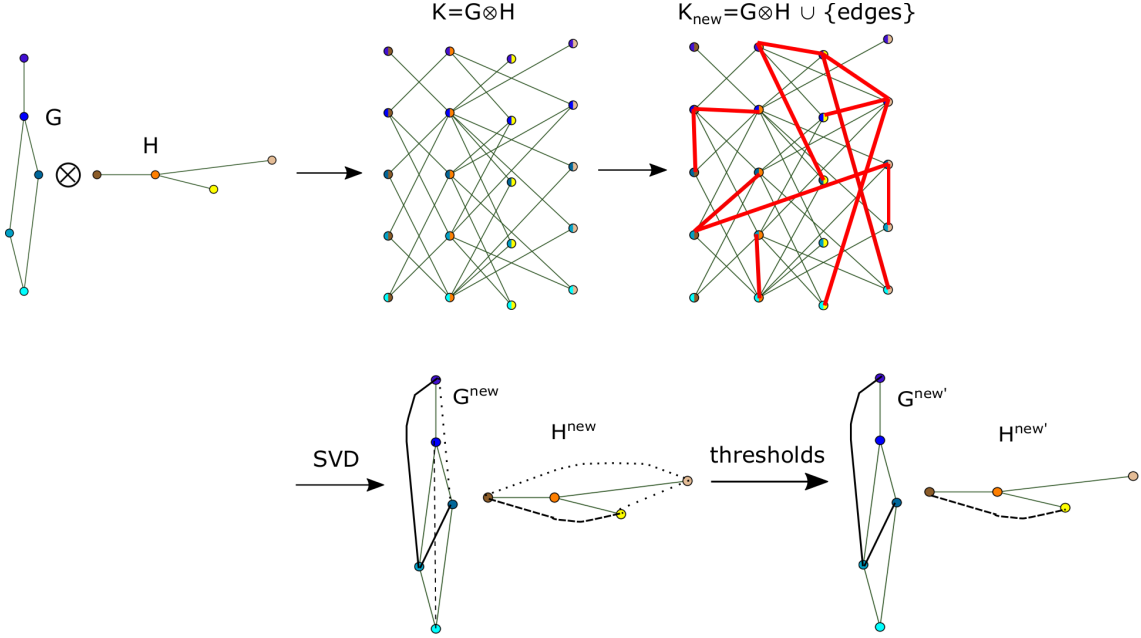}
\caption{The process of obtaining factor graphs from the non-Kronecker graph.}
\label{fig:procedure_steps}
\end{figure}

Let $G$ and $H$ be the graphs with the orders $n_{1}$ and $n_{2}$, respectively. The similarity matrix $S_{1}=(s_{ij}^{(1)})$, $i,j=1,...,n_{1}$ of $G$ is obtained by attaching the following weights to the edges $(i,j)$

\begin{equation}
\label{eq:weight}
\omega(i,j) =\left\{
\begin{array}{rl}
e^{-(y_{1i}^{'}-y_{1j}^{'})}, & (i,j)\in E(G)\\
0, & (i,j)\notin E(G)
\end{array},
\right.
\end{equation}


\noindent where we add random noise to the vector $y_{1}$ i. e. $y_{1}^{'} = y_{1} + \nu_{2}$, $\nu_{2}\in \mathcal{N}(0, 0.25)$. The same holds for the matrix $S_{2}$ and vector $y_{2}$. Therefore the similarity matrix of the Kronecker product $K = G\otimes H$ is calculated as $S(K) = S_{1}\otimes S_{2}$. For a violation of the Kronecker graph structure, new edges in the graph $K$ are added randomly (to obtain the graph $K_{new}$), by replacing selected 0-positions $(i,j)$ in the matrix $S(K)$ with the value $e^{-(y_{1i_{1}}^{'}-y_{1j_{1}}^{'})}e^{-(y_{1i_{2}}^{'}-y_{1j_{2}}^{'})}$, where $i_{1} = \lfloor (i-1)/n_2\rfloor + 1$, $j_{1}=(i-1)\% n_{2}+1$, $i_{2}=\lfloor (j-1)/n_{2}\rfloor + 1$ and $j_{2}=(j-1)\% n_{2} + 1$. We can notice that $S(K)_{i,j}=0$ if and only if $\omega_{i_{1}j_{1}}=0$ or $\omega_{i_{2}j_{2}}=0$ (in other words $(i_1,j_1)\notin E(G)$ or $(i_2,j_2)\notin E(H)$). In this way we obtain the so called near Kronecker graph $K_{new}$ (see Figure~\ref{fig:procedure_steps}), which means in general case that we cannot claim that $K_{new}$ can be represented as the Kronecker product of graphs.

After applying Corollary~\ref{cor:svd} on the similarity matrix $S(K_{new})$, we get two matrices $S_{1}^{new}$ and $S_{2}^{new}$ which are the similarity matrices of some graphs $G^{new}$ and $H^{new}$ (see Figure~\ref{fig:procedure_steps}). Also, we may notice that these matrices are non-negative and symmetric matrices which follows from Theorems 6 and 9 from \cite{van2000ubiquitous}, but some diagonal entries are nonzero. According to experimental results based on the edge density levels of graphs, the appropriate values are used as thresholds to remove quite a lot weak edges (this part is explained in more details in the following paragraph). Since matrices $S_{1}^{new}$ and $S_{2}^{new}$ are symmetric matrices, the newly obtained sparser matrices $S_{1}^{new'}$ and $S_{2}^{new'}$ are also symmetric. In order to get simple graphs, a rest of the non-zero diagonal elements (there are only a few left) are set to value 0. Therefore, the elements of $S_{k}^{new'}$ are defined as follows

\begin{equation}
\notag
S_{k}^{new'}(i,j) =\left\{
\begin{array}{rl}
S_{k}^{new}(i,j), & S_{k}^{new}(i,j) \geq t_{k}(\rho_{k})\\
0, & S_{k}^{new}(i,j) < t_{k}(\rho_{k})\; \text{or}\; i=j
\end{array}
\right. \text{,}\; k=1,2,
\end{equation}

\noindent where $t_{1}(\rho_{1})$ and $t_{2}(\rho_{2})$ are the thresholds, while $\rho_1$ and $\rho_2$ are the edge density levels of the initial graphs $G$ and $H$. Furthermore, a process of generation of unstructured predictor $R_{k}(x)\,\, (k = 1)$ is done according to the equations \eqref{eq:covariance_matrix} and \eqref{eq:mean}. Also, the parameters $\alpha$ and $\beta$ are set to the values 1 and 5, respectively.

Before we present the final results and the models accuracy, we give some comments related to the weights of the existing edges in graphs $G^{new}$ and $H^{new}$, after the nearest Kronecker product of $S(K_{new})$ is determined. At the beginning, it is important to mention that the graphs $G$ and $H$ are the subgraphs of $G^{new}$ and $H^{new}$, respectively. This implies that the edge density levels for both graphs $G^{new}$ and $H^{new}$ are higher than the edge density level of $G$ and $H$. After the approximation is applied, there are very small variations of the edge weights comparing only the weights of the edges which exist in both initial and new graphs. However, as much as the number of added edges in the graph $K$ is larger, these variations become much noticeable. Moreover, many of these new edges (which do not exist in initial graphs) are weak edges, which means that their weights are much less than the average edge weights (sometimes $10^{3}$ times less, even more). In our experiments these edges are treated as a noise and they can be easily removed by using certain threshold as a function of the values of similarity matrix. The thresholds $t_{1}=t_{1}(\rho_{1})$ and $t_{2}=t_{2}(\rho_{2})$ are determined by calculating percentiles for the matrices elements for both matrices $S_{1}^{new'}$ and $S_{2}^{new'}$.

\subsection{Results}

In performed experiments, where the model fitness is tested, the edge density level $\rho\in \{10\%, 20\%, 30\%\}$ is varied simultaneously for both graphs, $G$ and $H$, which are random networks of the same type. In the following text we present the obtained results for the regression accuracy of the GCRF models, when the previously described approximation (singular value decomposition) is applied to the near Kronecker graph $K_{new}$. Similarly as before, the comparison is done between five models: \textit{GCRF-base}, \textit{GCRF-baseSVD}, \textit{GCRF-MSN}, \textit{GCRF-LaplaceVec} and \textit{GCRF-NormLaplaceVec}, which are briefly described below. The GCRF-base model is based on the similarity matrix $S(K_{new})$ and numerical calculations for the matrix eigendecomposition. The GCRF-baseSVD model is based on the SVD approximation of the matrix $S(K_{new})$. After the matrix is approximated with the Kronecker product of matrices, numerical calculations for the matrix eigendecomposition are performed. In opposite to the GCRF-baseSVD and GCRF-base models, the approximations for the Laplacian eigenvalues and eigenvectors are used in the GCRF-MSN, GCRF-LaplaceVec and GCRF-NormLaplaceVec models as it was shown in the previous section, instead of the numerical calculations. Here, we present the results for experiments where the graphs are Erd\H{o}s-R\'enyi networks with 30 and 50 vertices. The same experiments were also conducted on the Erd\H{o}s-R\'enyi networks with 50 and 100 vertices, as well as with 100 and 200 vertices, and the results are pretty much the same.

\begin{figure}
\centering
\includegraphics[width=0.7\textwidth]{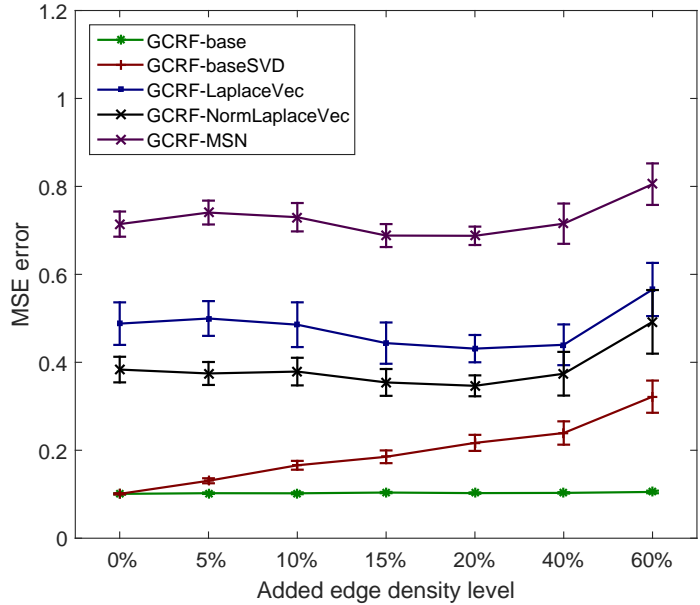}
\caption{Accuracy (MSE) of the models when the edge density levels of the initial graphs are 10\%.}
\label{fig:svd_10_10}
\end{figure}

In order to see how the percentage of added edges in the network influences the regression accuracy of the models, we conduct experiments where the number of added edges is linearly dependent on the number of existing edges in the network. The percentage of added edges is varied from the set $noise = \{0\%, 5\%, 10\%, 15\%, 20\%, 40\%, 60\%\}$ with respect to the number of edges in the initial Kronecker graph $K$. Here, we describe the experiments for the Erd\H{o}s-R\'enyi networks with 30 and 50 vertices, with edge density levels of 10\%, for both graphs (Figure~\ref{fig:svd_10_10}). When there are no additional edges in the graph $K$ (0\% of added edges), the results are the same as in Figure~\ref{fig:er_30x50_density_noise} for 10\% on $x$-axes. In this case the MSE errors obviously stem only from the estimation of Laplacian spectra of the Kronecker product of graphs ($K_{new}$ is the Kronecker product of graphs). Figure~\ref{fig:svd_10_10} represents the MSEs of the models when the percentage of added edges takes values from the set $noise$. It can be seen that the MSEs of the GCRF-base and GCRF-baseSVD models are very close to each other at the beginning, while their MSE difference becomes more noticeable when a percentage of added edges increases (differences between MSEs stems from the SVD approximation). Indeed, it can be noticed from Table~\ref{tbl:frobenius_varying} that the Frobenius norm of the difference between the similarity matrix $S(K_{new})$ and approximation with the Kronecker product is increasing when the number of added edges in the graph $K$ increases. Furthermore, when the number of added edges increases, the MSE gap between the approximation models (GCRF-base, GCRF-LaplaceVec, GCRF-NormLaplaceVec and GCRF-MSN) and GCRF-baseSVD models becomes smaller (noise from 10\% to 60\%). This can be explained with the fact that the eigenvectors used in these models become much better for a denser network. In this case, the error arising from the SVD approximation is a bit compensated with more stable estimations of the Laplacian eigenvalues and eigenvectors for a denser graph. Also, from 40\% to 60\% the MSEs graphics of all approximation models become a bit steeper than in the previous steps making the MSE gap between them and GCRF-base model larger. In this case the number of added edges is large, which leads to a serious violation of the Kronecker structure. A similar behavior is noticed when the initial graphs, $G$ and $H$ ($|G|=30$ and $|H|=50$), have edge density levels of 10\% and 20\%, 10 \% and 30\% and vice-versa.

\begin{table}
  \footnotesize
  \centering
  \setlength{\tabcolsep}{0.5em}
  \renewcommand{\arraystretch}{1.2}
  \begin{tabular}{|c|c|c|}
    \hline
     \textbf{noise} & \textbf{Frobenius norm 1} & {\textbf{Frobenius norm 2}}\\
    \hline
    5\%  & 10.607 & 10.601  \\ \hline
    10\% & 15.033 & 15.027  \\ \hline
    15\% & 18.818 & 18.811  \\ \hline
    20\% & 21.557 & 21.551  \\ \hline
    40\% & 30.513 & 30.501  \\ \hline
    60\% & 37.485 & 37.471  \\ \hline
  \end{tabular}
  \caption{$\left. a \right)\; \|S(K_{new})-S_{1}^{new}\otimes S_{2}^{new}\|_{F}$ and $\left. b \right) \; \|S(K_{new})-S_{1}^{new'}\otimes S_{2}^{new'}\|_{F}$, where edge density level for both initial graphs is 10\%.}
  \label{tbl:frobenius_varying}
\end{table}

\subsection{GCRF execution time with consecutive approximations}
\label{subsec:svd_time}

Here, we also check the execution time of each GCRF model, separately. Table~\ref{tbl:speed_svd} shows the results when two networks are Erd\H{o}s-R\'enyi networks with the edge density level of 30\%. The baseline algorithm (GCRF-base) has large computational complexity of $O(n_{1}^{3}n_{2}^{3})$, and so it is slower in respect with the GCRF models with approximations (GCRF-MSN, GCRF-NormLaplaceVec, GCRF-LaplaceVec). Compared to the results from Subsection~\ref{subsec:exec_time}, the approximation models, considered throughout this section, have singular value decomposition as additional approximation step, since the initial adjacency matrix has to be decomposed into the Kronecker product of two matrices. This step slows down the entire approximation models additionally, but as it can be noticed they are still much faster than the GCRF-base model. This follows from the fact that only the largest singular value and the corresponding eigenvectors have to be determined, not the complete singular value decomposition of a matrix. Unlike the previous, for the GCRF-base model all eigenvalues and eigenvectors have to be calculated. We omitted the execution time of the GCRF-baseSVD model, because it is given only for the purpose of models comparison, without practical usefulness.

\begin{table}
  \footnotesize
  \centering
  \setlength{\tabcolsep}{0.5em}
  \renewcommand{\arraystretch}{1.2}
  \begin{tabular}{|c|c|c|c|c|c|}
    \hline
      \multicolumn{2}{|c|}{\textbf{\#vertices}} & \multirow{2}{*}{\textbf{GCRF-base}} & \multirow{2}{*}{\textbf{GCRF-LaplaceVec}} & \multirow{2}{*}{\textbf{GCRF-NormLaplaceVec}} & \multirow{2}{*}{\textbf{GCRF-MSN}} \\
    \cline{1-2}
    \textbf{G} & \textbf{H} & & & & \\
    \hline
    50 & 30 & 0.78 $\pm$ 0.03 & 0.55 $\pm$ 0.04 & 0.71 $\pm$ 0.01 & 0.71 $\pm$ 0.01 \\ \hline
    50 & 100 & 19.03 $\pm$ 0.09 & 3.26 $\pm$ 0.03 & 3.41 $\pm$ 0.02 & 3.40 $\pm$ 0.02 \\ \hline
    100 & 200 & 1089 $\pm$ 8.99 & 67.32 $\pm$ 0.82 & 67.18 $\pm$ 0.59 & 67.45 $\pm$ 0.74 \\ \hline
    100 & 300 & 2216 $\pm$ 6.21 & 160.37 $\pm$ 6.28 & 159.6 $\pm$ 7.18 & 159.21 $\pm$ 5.44 \\ \hline
    200 & 200 & 5192 $\pm$ 43.6 & 349.98 $\pm$ 21.66 & 348.69 $\pm$ 28.30 & 345.33 $\pm$ 25.56 \\ \hline
  \end{tabular}
  \caption{Execution time in seconds for the Kronecker product of the Erd\H{o}s-R\'enyi networks with the edge density level of 30\%.}
  \label{tbl:speed_svd}
\end{table}

\section{Conclusion}

As it was theoretically and experimentally shown, the GCRF model has high computational complexity and as such it is non-scalable for large networks with tens of thousands nodes. In last ten years, a few approaches, more or less successful, are developed in order to reduce the running time of GCRF on large networks and to preserve high prediction accuracy. The fact that the whole network could be represented as the Kronecker product of graphs was used as a possibility for speeding up the GCRF learning task [Glass and Obradovic, 2017]. However, approximations for the eigenvalues and eigenvectors which are used in this model are not suitable, which is reflected in high GCRF MSE, since characterizing a Laplacian spectrum of such a graph from spectra of its factor graphs has remained open and challenging problem. In this paper we apply new estimations of the Laplacian eigenvalues and the corresponding eigenvectors for the Kronecker product of graphs in the GCRF model. A computational complexity of these approximations is much less than that of explicit computation of eigenvalues of a product graph. To evaluate the proposed models, we conducted experiments on three type of random networks: Erd\H{o}s-R\'enyi, Watts-Strogatz and Barab\'asi-Albert networks. A significant accuracy improvement is achieved compared to the GCRF model used in [Jesse and Obradovic]: when the initial networks are Erd\H{o}s-R\'enyi random networks, the obtained MSEs of the proposed approximation models are more than 3 times lower than the MSE of the previously proposed model, and more than 2 times in the case of Barab\'asi-Albert networks. Also, it was shown that the GCRF model which incorporates the estimated eigenvalues and eigenvectors from Subsection~\ref{subsec:sayama_approx} achieved good prediction accuracy in the case of Barab\'asi-Albert random networks, while the approximations from Subsection~\ref{subsec:novel_approx} are more suitable in the case of Erd\H{o}s-R\'enyi and Watts-Strogatz random networks (the GCRF-NormLaplaceVec error tends to the MSE of the GCRF-base model).

The same models are also tested in the case when the network factorization into a Kronecker product of networks is not possible. In this case, the Singular Value Decomposition (SVD) is used for finding the nearest Kronecker product of graphs for an initial graph. A combination of two consecutively applied approximations, SVD and approximations for the Laplacian eigenvalues and eigenvectors, provide high regression accuracy of the approximated GCRF models. Although there are two approximations involved in the GCRF model, the execution time of such models is much shorter compared to the execution time of original GCRF model, while the achieved MSE is low. In the future research we will try to approximate the non-Kronecker graph with as much as possible value of the rank, that is, with the corresponding permuted matrices with the ranks higher than one (according to Theorem~\ref{thm:approx}).

Take into consideration that the experimental setup is designed when the edge density levels of initial graphs are given in advance. This information is used during the weak edges removal stage. For the future work it will be good to see how the model accuracy is changing in a situation when the initial edge density levels of graphs are not given in advance, but we approximately know their values according to problem domain knowledge. Then, after weak edges removal process, the edge density levels of graphs could be a bit larger or smaller.


\bibliography{references}

\appendix

\section{}
\label{app:gcrf_small_graphs}

\begin{figure}[!htbp]
\centering
\includegraphics[width=\textwidth]{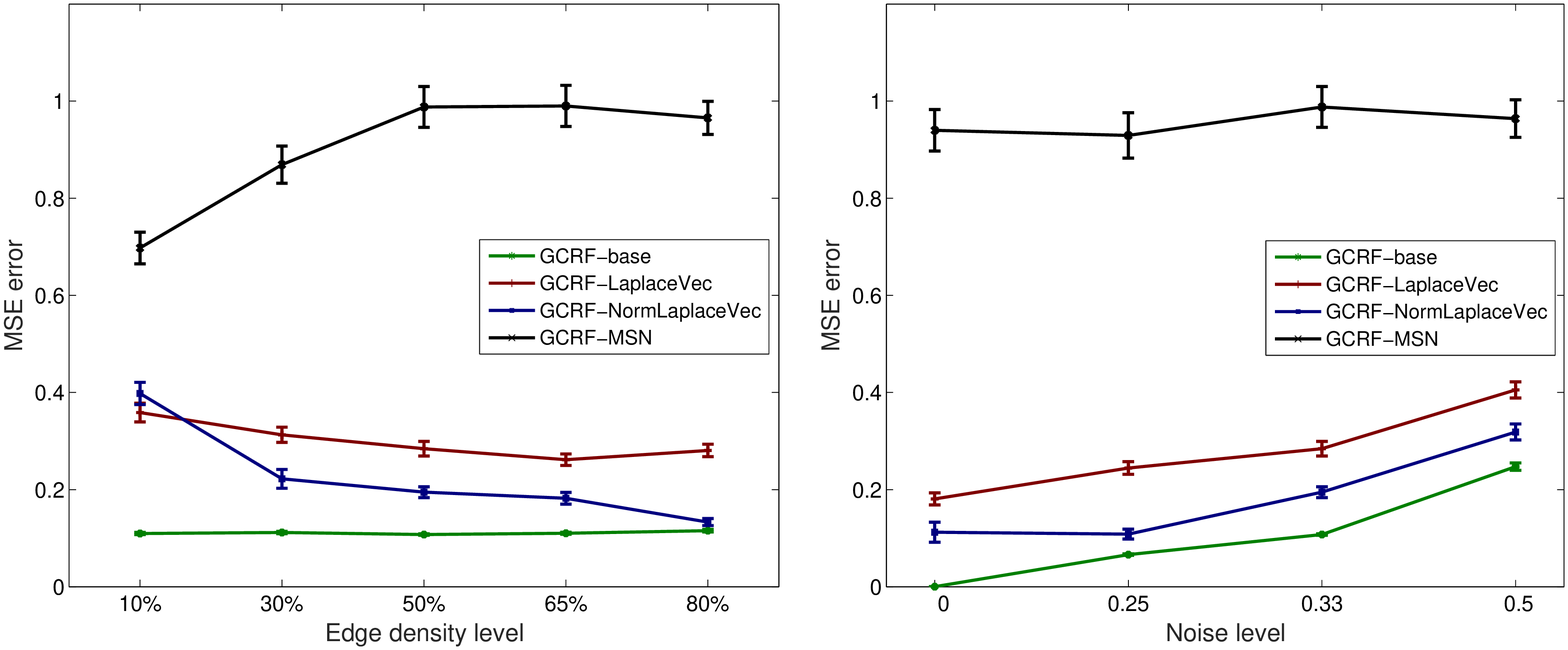}
\caption{Accuracy (MSE) of the models as a function of edge density percentage (left) and noise level (right), where $S_{1}$ = Erd\H{o}s-R\'enyi random graph with 30 vertices and $S_{2}$ = Erd\H{o}s-R\'enyi random graph with 50 vertices. \textit{Left:} corresponding number of edges are \{43, 130, 217, 282, 348\} and  \{122, 367, 612, 796, 980\}. \textit{Right:} accuracy (MSE) of the models with respect to different noise, for fixed 50\% edge density percentage.}
\label{fig:er_30x50_density_noise}
\end{figure}

\begin{figure}[!htbp]
\centering
\includegraphics[width=\textwidth]{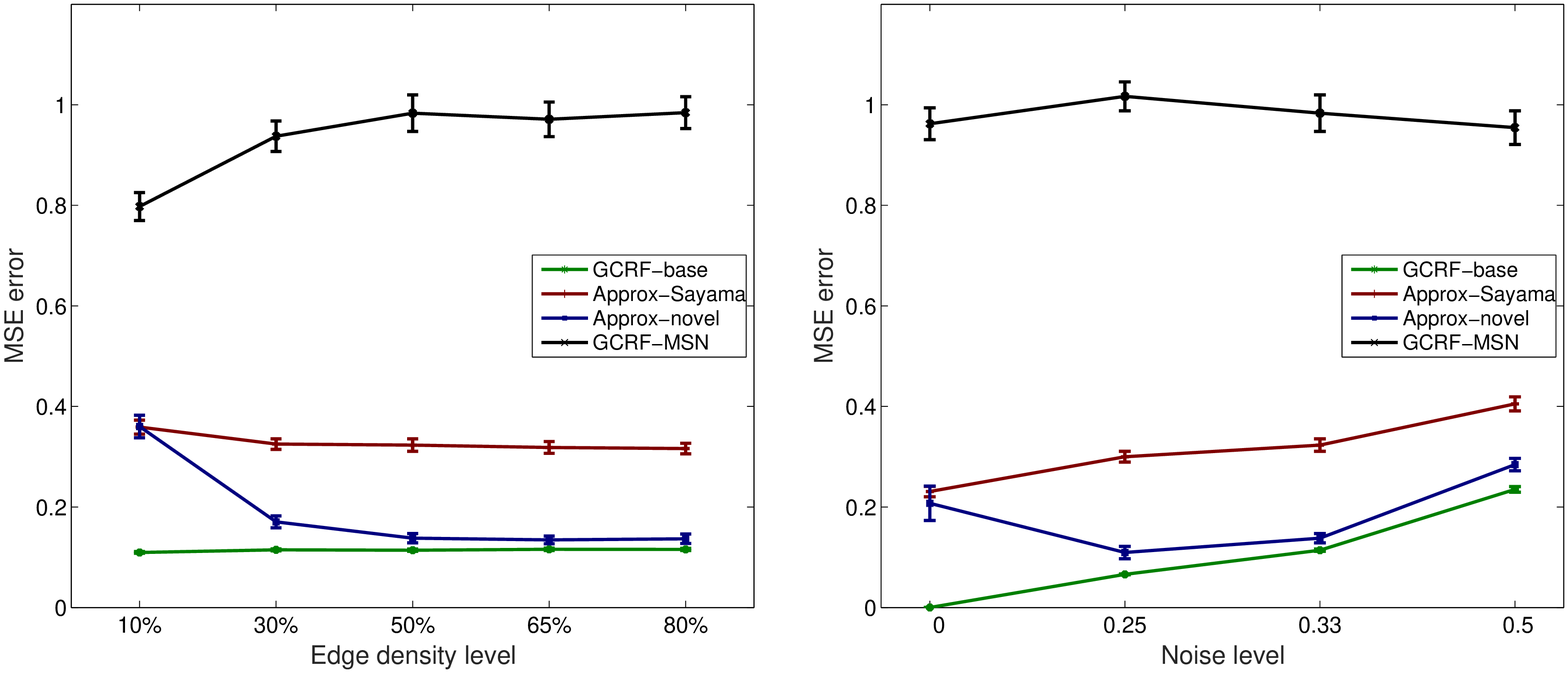}
\caption{Accuracy (MSE) of the models as a function of edge density percentage (left) and noise level (right), where $S_{1}$ = Erd\H{o}s-R\'enyi random graph with 50 vertices and $S_{2}$ = Erd\H{o}s-R\'enyi random graph with 100 vertices. \textit{Left:} corresponding number of edges are \{122, 367, 612, 796, 980\} and  \{495, 1485, 2475, 3217, 3980\}. \textit{Right:} accuracy (MSE) of the models with respect to different noise, for fixed 50\% edge density percentage.}
\label{fig:er_50x100_density_noise}
\end{figure}

\begin{figure}[!htbp]
\centering
\includegraphics[width=\textwidth]{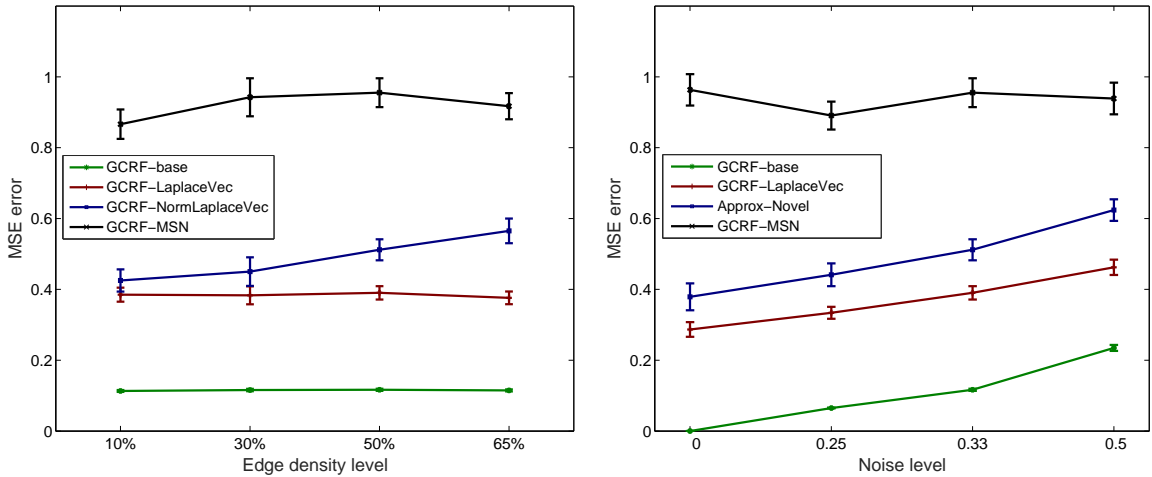}
\caption{Accuracy (MSE) of the models as a function of edge density percentage (left) and noise level (right), where $S_{1}$ = Barab\'asi-Albert random graph with 30 vertices and $S_{2}$ = Barab\'asi-Albert random graph with 50 vertices. \textit{Left:} corresponding number of edges are \{56, 125, 216, 225\} and \{141, 369, 616, 625\}. \textit{Right:} accuracy (MSE) of the models with respect to different noise, for fixed 50\% edge density percentage.}
\label{fig:ba_30x50_density_noise}
\end{figure}

\begin{figure}[!htbp]
\centering
\includegraphics[width=\textwidth]{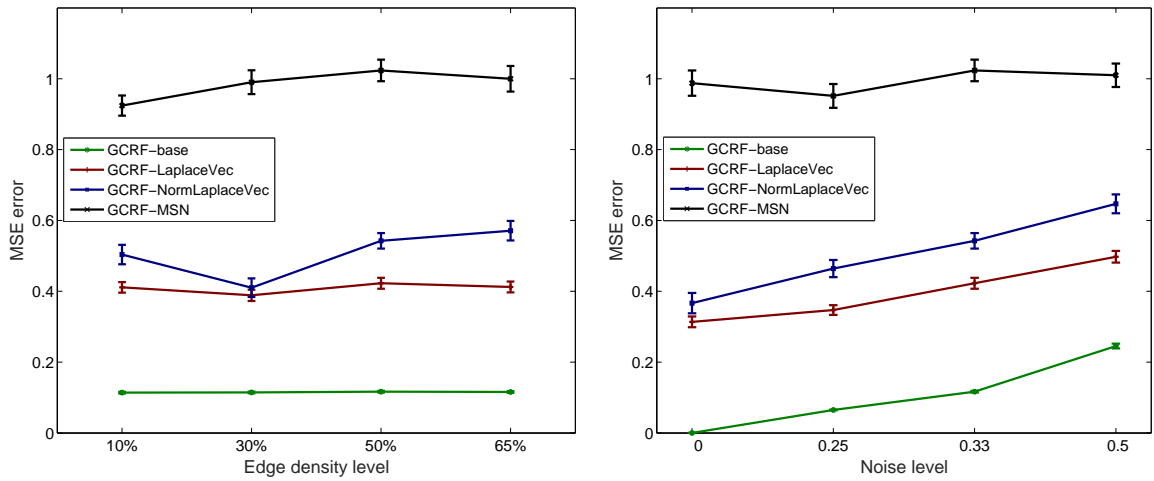}
\caption{Accuracy (MSE) of the models as a function of edge density percentage (left) and noise level (right), where $S_{1}$ = Barab\'asi-Albert random graph with 50 vertices and $S_{2}$ = Barab\'asi-Albert random graph with 100 vertices. \textit{Left:} corresponding number of edges are \{141, 369, 616, 625\} and \{475, 1476, 2475, 2500\}. \textit{Right:} accuracy (MSE) of the models with respect to different noise, for fixed 50\% edge density percentage.}
\label{fig:ba_50x100_density_noise}
\end{figure}

\end{document}